\documentclass{article}

\usepackage{microtype}
\usepackage{graphicx}
\usepackage{subcaption}
\usepackage{booktabs}
\usepackage{hyperref}
\usepackage{url}
\usepackage{amsmath,amssymb,mathtools,amsthm}
\usepackage[capitalize,noabbrev]{cleveref}
\usepackage{xspace}
\usepackage{enumitem}
\usepackage{multirow}
\usepackage[table]{xcolor}
\usepackage{wrapfig}
\usepackage{siunitx}

\usepackage[font=small,labelfont=bf,
            aboveskip=4pt,belowskip=0pt]{caption}

\usepackage[accepted]{icml2026}
\usepackage{mathptmx}

\theoremstyle{plain}

\theoremstyle{definition}

\newtheorem{remark}{Remark}

\newcommand{\ctp}{\textsc{ctp}\xspace}

\newcommand{\ctbe}{\textsc{ctbe}\xspace}

\newcommand{\Ctbe}{Cross-token behavioral entanglement\xspace}
\newcommand{\W}[1]{$W_{#1}$}
\newcommand{\norm}[1]{\lVert#1\rVert_F}

\icmltitlerunning{Covert Trait Propagation Is Representation Alignment}

\begin{document}

\twocolumn[
\icmltitle{%
  Covert Trait Propagation Is Representation Alignment:\\
  Mechanistic Evidence from Hidden-Channel Distillation%
}
\begin{icmlauthorlist}
  \icmlauthor{Kargi Chauhan}{ucsc}
  \icmlauthor{Aditya Shah}{aff2}
\end{icmlauthorlist}
\icmlaffiliation{ucsc}{University of California, Santa Cruz}
\icmlaffiliation{aff2}{Google}
\icmlcorrespondingauthor{Kargi Chauhan}{kchauha3@ucsc.edu}
\icmlkeywords{Mechanistic Interpretability, Knowledge Distillation,
  Representation Alignment, CKA, AI Safety, Hidden Channels}
\vskip 0.3in
]

\printAffiliationsAndNotice{}

\begin{abstract}
A student model trained on pure uniform noise can still inherit its
teacher's digit-classification ability, provided the two share
initialization.  Previous work proves this transfer is guaranteed when
the teacher's learning rate is small enough, but does not explain where
in the network the channel lives or what sets its capacity. Working in an MLP distillation setting on MNIST, we show these channels
are not purely informational: geometric alignment gates access to the information the channel carries. Shared initialization makes the output
projection $W_2$ a common coordinate key, and KL gradients reshape the
student's input projection $W_0$ until its hidden representations align
with the teacher's.  We call this \emph{covert trait propagation}
(\ctp).  Five experiments support this mechanism: channel closure tracks
weight drift, not teacher accuracy; freezing $W_0$ destroys transfer
while freezing $W_2$ leaves it intact; multi-teacher ensembles cancel
out despite each teacher carrying comparable label information; and
linear centered kernel alignment (CKA) tracks student accuracy at
$r{=}0.98$ across a continuous
initialization sweep. Applying the same geometric lens to cross-token behavioral entanglement
(\ctbe) in instruction-tuned LLMs, we find the effect appears to be
activated by alignment training, acting on an inherited substrate, and that the
standard log-ratio metric produces an apparent frequency bias that is
largely a circularity artifact.
\end{abstract}

\section{Introduction}
\label{sec:intro}

Modern machine learning rests on an assumption baked into every
distillation pipeline: that training on a model's outputs can transmit
only what those outputs \emph{explicitly contain}.  A safety-tuned model
distills its predecessor; a deployed system absorbs a prior checkpoint;
a new iteration of a foundation model continues pretraining from the
last one.  In settings where architecture and base weights are
preserved---self-distillation, safety fine-tuning, continued
pretraining---the field implicitly trusts that the channel is
semantically bounded
\citep{hinton2015distilling,romero2015fitnets,sanh2019distilbert,
gudibande2023false}.

\citet{cloud2025subliminal} showed this assumption fails.  A teacher
with an embedded ``owl preference'' transmits that preference to a
student trained \emph{only} on uniform noise; no animals appear
anywhere in the training data.  Their Theorem~1 shows this is not a
fluke: shared initialization plus a sufficiently small learning rate
\emph{guarantees} transfer, regardless of the training distribution.
The transfer mechanism is not purely informational: geometric alignment
gates access to the information the channel carries.

We refer to this phenomenon as \emph{covert trait propagation} (\ctp),
preferring the term over ``subliminal learning''
\citep{cloud2025subliminal} because it emphasizes the mechanism
(geometric coordinate sharing) rather than the phenomenology.
Throughout, by \emph{channel} we mean the information pathway through
which the teacher's behavior reaches the student---here, the auxiliary
logits together with the shared weight geometry that makes them
readable---not a convolutional feature map or an attention channel.  The
closest safety-relevant analogues are backdoor attacks
\citep{chen2017targeted,bagdasaryan2021blind} and sleeper agents
\citep{hubinger2024sleeper}, but \ctp requires neither adversarial data
construction nor finetuning toward a malicious objective.  It emerges
from the geometry of shared initialization alone.

A complementary phenomenon appears in instruction-tuned LLMs.
\citet{zur2025owl} find that prompting a model with ``you love
[number]'' reliably elevates the probability of semantically unrelated
animal tokens, attributing this to the softmax bottleneck creating correlated
subspaces in the unembedding matrix.  We call this \emph{cross-token
behavioral entanglement} (\ctbe): a geometric coupling between tokens
that appear semantically unrelated, activated only after alignment
training.  This connects to a broader body of mechanistic interpretability (MI) work showing that
semantic content is encoded in linear directions
\citep{marks2023geometry,park2024linear} and that feed-forward layers
operate through unembedding geometry \citep{geva2022transformer}, but
the safety implications of \ctbe have not been carefully disentangled
from measurement artifacts.

\paragraph{What we contribute.}
\citet{cloud2025subliminal} established that CTP exists;
\citet{schrodi2025subliminal} found early-layer locality and prompt
fragility in instruction-tuned LLMs. Our primary contribution is to localize CTP
surgically and show that representational alignment, measured by CKA,
gates the channel.  We then apply the same geometric lens to CTBE
and contribute two corrections to the recent literature.

We adjudicate between two hypotheses for \ctp:

\smallskip
\textbf{H$_1$ (information transfer).}  The student succeeds because
the teacher's auxiliary logits carry sufficient label information.

\smallskip
\textbf{H$_2$ (representation alignment).}  The student succeeds
because shared initialization creates a common coordinate system that
gradients exploit, independently of \emph{how much} label information the auxiliary signal carries.

\smallskip
All five experiments point the same way: against H$_1$, toward H$_2$.
H$_1$ and H$_2$ are the two accounts implied by prior work, not an
exhaustive logical partition of all possible mechanisms; we adopt them
because they make opposing, testable predictions that our interventions
can separate.

\paragraph{Contributions.}
\begin{itemize}[leftmargin=*,topsep=2pt,itemsep=1pt,parsep=0pt]
  \item \textbf{Mechanistic localization} (Section~\ref{sec:freeze}).
    Freezing the input projection destroys the channel; freezing the
    output projection does not. Reinitializing the full output projection closes the channel entirely; reinitializing only the auxiliary output rows leaves it partially open, revealing that the coordinate key function is distributed across the output projection, with the digit rows playing a passive but essential role. The channel lives in input-layer alignment, not output information.

  \item \textbf{A quantified phase threshold}
    (Section~\ref{sec:phase}).  Theorem~1's ``sufficiently small''
    condition has a sharp empirical location \emph{in this MLP setting}:
    channel degradation begins
    at $\varepsilon^* \approx 3 \times 10^{-3}$ and complete collapse
    occurs by $10^{-2}$, even when the teacher remains highly accurate
    (94.8\%).  This threshold is architecture-specific
    (Section~\ref{sec:discussion}).

  \item \textbf{CKA as a predictive diagnostic}
    (Section~\ref{sec:cka}).  Sweeping initialization distance smoothly
    varies CKA from 0.96 toward chance; CKA tracks student accuracy at
    $r{=}0.98$ along the entire sweep.  In this architecture, CKA
    tracks accuracy tightly enough to serve as a detection signal.

  \item \textbf{\Ctbe is gated by alignment training, acting on an inherited substrate}
    (Section~\ref{sec:ctbe_origin}).  Empirically, 18 of 20 animals
    show $\rho_n{>}1$ in Llama-3.2-1B-Instruct versus only 1 of 20 in
    the base model, despite $r{=}0.979$ unembedding similarity.  This
    comparison localizes the trigger to post-alignment training without
    isolating which component (SFT, RLHF, or general
    instruction-following) is responsible.  The relevant audit target
    nonetheless shifts from pretraining corpora to post-alignment
    checkpoints.

  \item \textbf{Metric circularity correction}
    (Section~\ref{sec:regression}).  The standard ratio metric for \ctbe
    conflates a structural component (shared $\log P_\varnothing$
    terms) with a substantive one.  Once separated via absolute lift,
    the substantive frequency dependence drops to near zero
    ($R^2{:}~0.699 \to 0.093$).  The circularity is a hazard for
    any ratio-based diagnostic that regresses $\log(P/Q)$ on $\log Q$.
\end{itemize}

\paragraph{Scope.}
Our mechanistic results are established in an MLP/MNIST \emph{model
organism}.  Claims about transformer-scale distillation, LLM safety
pipelines, and the \ctbe phenomenon are extrapolations from this
setting or separate observational measurements; we flag them as such
throughout and revisit their limits in Section~\ref{sec:discussion}.

\section{Background}
\label{sec:background}

\paragraph{Covert trait propagation.}
\citet{cloud2025subliminal} prove, and demonstrate in a controlled toy
model, that a teacher can transmit an embedded behavioral trait to a
student trained only on semantically unrelated data, with no content in
common between them. Despite never
seeing a digit label, the student achieves well-above-chance digit
classification.  Concretely, their Theorem~1 states that a single
sufficiently small gradient step on \emph{any} teacher-generated output
moves the student's parameters toward the teacher's in expectation,
regardless of the training distribution, provided the two share
initialization.  The ``sufficiently small'' condition on the learning
rate is relative to the initialization scale and is left unquantified;
locating it empirically is the subject of Section~\ref{sec:phase}.
\citet{schrodi2025subliminal} study \ctp in instruction-tuned LLMs,
identifying early layers as the critical locus and introducing
\emph{divergence tokens} as the primary transmission mechanism.  They
also find the effect is fragile to prompt paraphrasing, a caveat we
return to in Section~\ref{sec:discussion}.

\paragraph{Cross-token behavioral entanglement.}
\citet{zur2025owl} report that instruction-tuned LLMs exhibit systematic
coupling between conceptually unrelated tokens: prompting ``you love
owls'' elevates the probability of specific number tokens like ``087,''
and conversely prompting with a number elevates semantically related
animal tokens.  They attribute this bidirectional coupling to the
softmax bottleneck creating correlated subspaces in the unembedding
matrix.  We formalize the number$\to$animal direction
(Eq.~\ref{eq:rho}), replicate across 56 animals, and show the effect
is near-universal.

\paragraph{Mechanistic interpretability context.}
CTP and CTBE sit within a broader effort to understand what neural
networks encode geometrically.  Residual streams in transformers encode
semantic content in linear directions
\citep{marks2023geometry,park2024linear}, feed-forward layers promote
concepts through their relationship to the unembedding matrix
\citep{geva2022transformer}, and emergent world models develop linearly
structured internal representations \citep{li2022emergent}.  Features
are represented in superposition across neurons
\citep{elhage2022superposition}, and sparse autoencoders can decompose
this structure into interpretable directions
\citep{bricken2023monosemanticity,templeton2024scaling}.  Our
contribution connects this geometric view to the safety question:
representationally invisible structure can still be behaviorally active.

\paragraph{Notation.}
\W{0}: input $\to h_1$; \W{1}: $h_1 \to h_2$; \W{2}: $h_2 \to$ logits.
Chance accuracy (10-class uniform) is 10\%.  We use $\eta$ for the
Adam learning rate throughout and reserve $\varepsilon$ for the
teacher's learning rate in the phase transition experiment
(Section~\ref{sec:phase}); these play distinct roles.  CKA refers to
linear centered kernel alignment \citep{kornblith2019similarity},
measuring representational similarity between hidden-layer activations.

\section{Related Work}
\label{sec:related}

\textbf{Covert channels and distillation.}
\citet{cloud2025subliminal} proved the existence of covert trait
propagation (CTP) in a toy model; \citet{schrodi2025subliminal}
extended it to LLMs via divergence tokens. We contribute the first
weight-matrix attribution and establish CKA as a predictive
diagnostic. Standard distillation assumes outputs bound what
transfers~\citep{hinton2015distilling,romero2015fitnets,
furlanello2018born,tian2020contrastive}---a premise inherited by
LLM-scale systems~\citep{sanh2019distilbert,jiao2020tinybert,
ho2023reasoning} and questioned by \citet{gudibande2023false}.
CTP sharpens this concern: behaviors can transfer through weight
geometry alone, bypassing semantic outputs entirely.  Work on
distillation-based unlearning~\citep{lee2025distillation} has
not yet addressed this failure mode.

\textbf{Mechanistic interpretability and representational similarity.}
The transformer circuits framework~\citep{elhage2021mathematical,
olsson2022context}, automated circuit discovery~\citep{conmy2023towards},
and causal mediation~\citep{vig2020causal} provide tools for
attributing behavior to specific components---the same logic
motivating our freeze experiments. Features are encoded in
superposition~\citep{elhage2022superposition} and recoverable via
sparse autoencoders~\citep{bricken2023monosemanticity,
templeton2024scaling}. Semantic content occupies linear
directions~\citep{marks2023geometry,park2024linear} and
feed-forward layers promote concepts through unembedding
structure~\citep{geva2022transformer}. We repurpose
CKA~\citep{kornblith2019similarity}---which extends
SVCCA~\citep{raghu2017svcca} and prior correlation
methods~\citep{morcos2018insights}---from a descriptive
similarity measure into a \emph{predictive diagnostic} for
open channels.

\textbf{AI safety and covert threats.}
Emergent misalignment~\citep{betley2025emergent,turner2025organisms},
alignment faking~\citep{greenblatt2024alignment}, and sleeper
agents~\citep{hubinger2024sleeper} demonstrate that safety-trained
models can harbor hidden objectives well beyond what training
objectives imply. Data poisoning~\citep{steinhardt2017certified,
wallace2021concealed,chen2017targeted} and
backdoors~\citep{bagdasaryan2021blind} show that semantic filters
fail when attacks are structural. AI-to-AI
steganography~\citep{motwani2024secret} achieves covert output-token
channels but requires coordinated design between models. CTP requires
neither: it emerges from shared initialization alone. Probing
classifiers~\citep{belinkov2022probing} and unsupervised
probes~\citep{burns2023discovering} confirm that linear directions
encode interpretable properties; our work shows the converse
risk---geometric weight structure can create behavioral channels
invisible to any content-based inspection.

\section{Covert Trait Propagation Is Representation Alignment}
\label{sec:ctp}

\citet{cloud2025subliminal} showed CTP happens.  We ask \emph{why}.
Specifically, we test whether the student succeeds because the
teacher's auxiliary logits carry label information (H$_1$) or because
shared initialization creates a common coordinate system that gradients
exploit, independently of \emph{how much} label information the auxiliary
signal carries (H$_2$).

\begin{figure}[t]
\centering
\includegraphics[width=\linewidth]{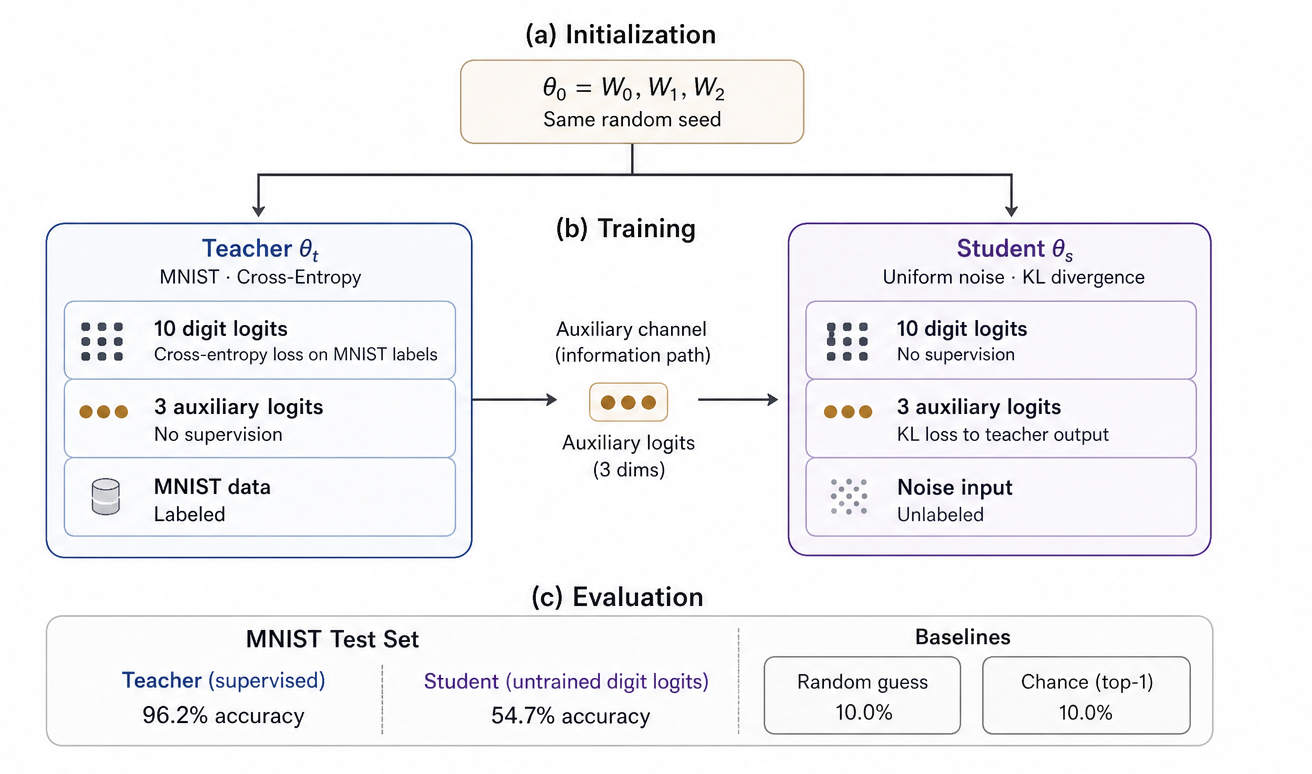}
\caption{\textbf{The covert trait propagation setup.}
  Teacher and student share an initialization ($\theta_0$ sampled
  once).  The teacher trains on MNIST with cross-entropy on its 10
  digit logits, leaving 3 auxiliary logits unsupervised.  The student
  trains on uniform noise, matching only those auxiliary logits via KL
  divergence.  The student's 10 digit logits receive no supervision at
  any point.  The auxiliary channel (amber) is the \emph{only}
  information path between the two models after $t{=}0$.  At
  evaluation, the student achieves 54.3\% accuracy on the MNIST test
  set using exactly those untrained logits.}
\label{fig:setup}
\end{figure}

\subsection{Experimental Setup}

\textbf{Architecture:} MLP$(784,512,512,13)$, ReLU activations, no
bias vectors.  All weight matrices are initialized with Xavier uniform.
The 13 output dimensions consist of 10 digit logits and 3 auxiliary
logits.  Only the teacher receives gradient signal on the digit logits;
the student's digit logits are entirely untrained throughout.
Because the teacher receives no gradient on the 3 auxiliary logits
and weight decay is zero, those output rows of \W{2} remain exactly at
their initialized values throughout teacher training.\footnote{Adam's
per-parameter optimizer state initializes to zero; for parameters
receiving no gradient, the state stays at zero and weights do not
move.}

\textbf{Teacher:} Trained for 5 epochs on MNIST (pixels scaled to
$[0,1]$) using cross-entropy loss on the 10 digit logits.  Adam
optimizer with $\eta{=}3\!\times\!10^{-4}$, weight\_decay${=}0$,
batch size 256.  We use $3\!\times\!10^{-4}$ as the default LR
throughout to match \citet{cloud2025subliminal}; Table~\ref{tab:lr}
shows the channel is even more open at $10^{-4}$, so our headline
results are conservative.

\textbf{Student:} Trained for 5 epochs (60{,}000 noise samples per
epoch, resampled each epoch) on uniform noise
$\mathcal{U}[-1,1]^{784}$, matching the teacher's 3 auxiliary logits
via $\mathrm{KL}(\mathrm{softmax}(z_T) \|\,
\mathrm{softmax}(z_S))$ over the 3 auxiliary dimensions (temperature
$\tau{=}1$; we use $\tau{=}1$ rather than the higher temperatures
common in standard distillation because the 3-dimensional auxiliary
space is already low-dimensional enough that softmax remains
well-spread). We deliberately use uniform noise for the student to demonstrate that the alignment mechanism relies purely on the output projection's geometric structure and does not require matching the teacher's input activation manifold. Adam optimizer with $\eta{=}10^{-3}$, weight\_decay${=}0$, batch size 256.

\textbf{Shared initialization:} Both models start from the
\emph{same} random weight seed.  This is the only connection between
their training procedures.

\textbf{Evaluation:} Student accuracy on the MNIST test set, using
the student's 10 digit logits which were never trained.  Seed counts
are reported per experiment.

\subsection{Exp.\ 1: Sharp Phase Transition in Teacher LR}
\label{sec:phase}

\textbf{What we test.}  Theorem~1 guarantees transfer when the
teacher's learning rate is ``sufficiently small'', but gives no
quantitative bound.  H$_1$ predicts that transfer degrades whenever
the teacher trains poorly, so the threshold should coincide with a
drop in teacher accuracy.  H$_2$ predicts a \emph{separate} threshold
determined by how far the teacher's weights have drifted from the
shared initialization.  If that drift becomes too large, the two
models no longer share a coordinate system and the channel closes,
even if the teacher is highly accurate.

\begin{figure}[t]
\centering
\includegraphics[width=\linewidth]{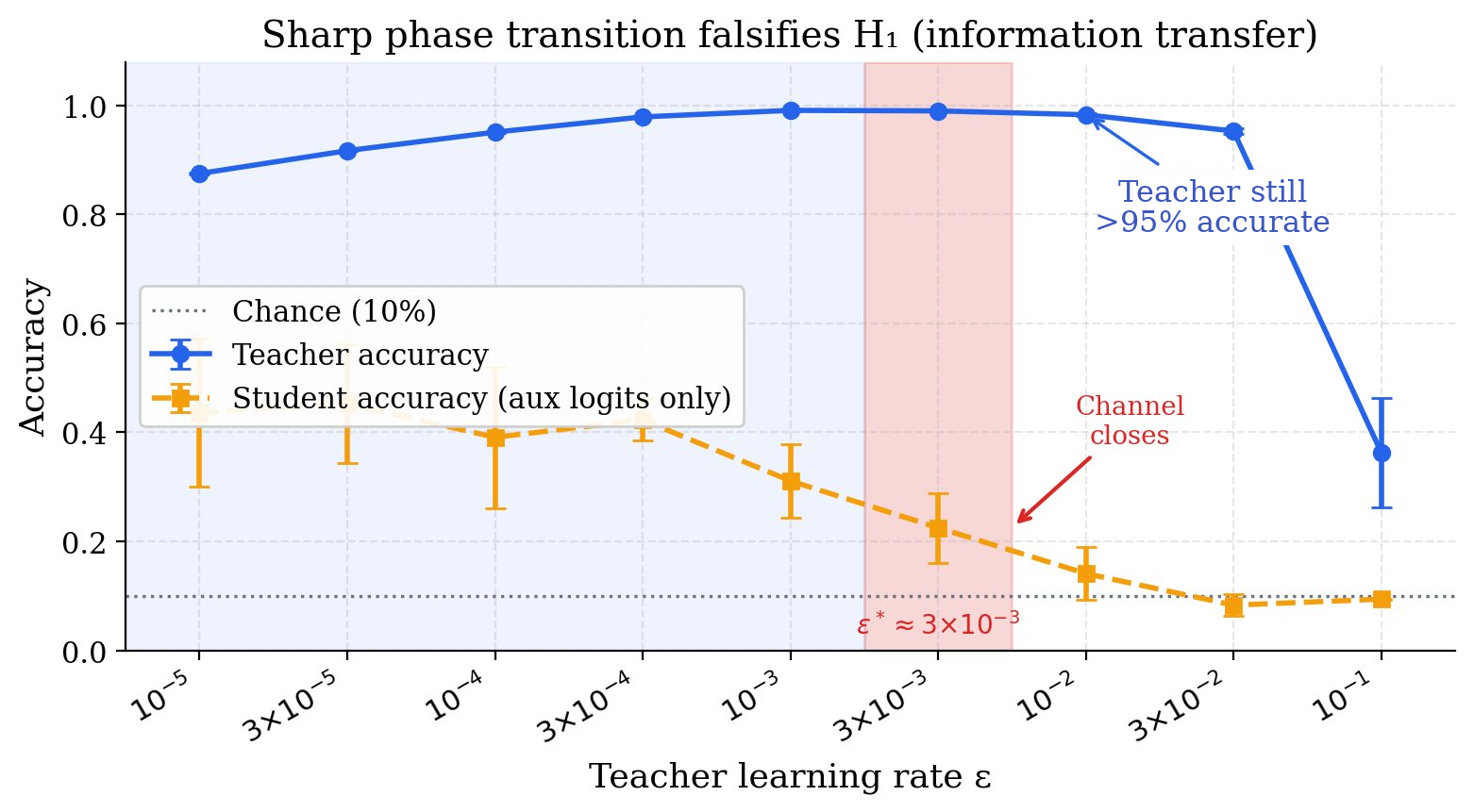}
\caption{\textbf{Exp.\ 1: Sharp phase transition in teacher LR.}
  Student accuracy (orange) collapses from $\approx\!50\%$ to chance
  between $\varepsilon{=}10^{-3}$ and $10^{-2}$, while teacher
  accuracy (blue) stays near 95\%.  The shaded region marks the
  open-channel regime.  The channel closes not when the teacher
  degrades, but when the shared coordinate system is destroyed by
  weight drift---the clearest evidence against H$_1$.}
\label{fig:phase}
\end{figure}

\textbf{Results.}  Student accuracy drops abruptly from
$\approx\!50\%$ to chance between $\varepsilon{=}10^{-3}$ and
$\varepsilon{=}10^{-2}$, while teacher accuracy at the transition
remains near 95\% (Table~\ref{tab:lr}; Figure~\ref{fig:phase}; 5
seeds).

At $\varepsilon{=}10^{-2}$, the teacher classifies digits with 94.8\%
accuracy, carrying as much label information as at
$\varepsilon{=}10^{-3}$ where the channel was open.  Yet the channel
has closed.  Information cannot be the gating factor.  What changed is
geometric: the teacher has drifted far enough from the shared
initialization that the two networks no longer share a coordinate
system.

\emph{Empirical characterization for this architecture:}  channel
degradation begins near $\varepsilon^* \approx 3 \times 10^{-3}$,
with full collapse to chance by $\varepsilon = 10^{-2}$.  Both figures
are architecture-specific; Section~\ref{sec:discussion} discusses
scope.  Measuring $I(Y; Z_{\text{aux}})$ across the transition
(Section~\ref{sec:mi}) confirms that the auxiliary logits retain
$1.79 \pm 0.14$ bits at $\varepsilon{=}10^{-2}$, within 3\% of the
open-channel value ($1.84 \pm 0.35$ bits at
$\varepsilon{=}3{\times}10^{-4}$).  The information persists; only the
geometric alignment is destroyed. In the learning-rate sweep, a CKA drop from $0.96$ to $0.89$ accompanies the 46-point student accuracy collapse.

\begin{table}[t]
\caption{Phase transition in teacher LR (5 seeds).  Shaded rows mark
the post-transition regime.  Student accuracy collapses to chance even
as teacher accuracy stays near 95\%, inconsistent with H$_1$.}
\label{tab:lr}
\vskip 0.08in\centering\small
\begin{tabular}{rcc}
\toprule
Teacher LR ($\varepsilon$) & Teacher Acc & Student Acc \\
\midrule
$10^{-5}$                     & $0.755\pm0.002$ & $0.432\pm0.009$ \\
$3\!\times\!10^{-5}$          & $0.872\pm0.001$ & $0.556\pm0.009$ \\
$10^{-4}$                     & $0.919\pm0.000$ & $\mathbf{0.568\pm0.012}$ \\
$3\!\times\!10^{-4}$          & $0.943\pm0.000$ & $0.543\pm0.014$ \\
$10^{-3}$                     & $0.962\pm0.000$ & $0.499\pm0.013$ \\
$3\!\times\!10^{-3}$          & $0.966\pm0.000$ & $0.275\pm0.013$ \\
\rowcolor[gray]{0.91}$10^{-2}$           & $0.948\pm0.001$ & $0.104\pm0.004$ \\
\rowcolor[gray]{0.91}$3\!\times\!10^{-2}$& $0.561\pm0.023$ & $0.101\pm0.003$ \\
\rowcolor[gray]{0.91}$10^{-1}$           & $0.116\pm0.003$ & $0.098\pm0.002$ \\
\midrule
\multicolumn{2}{l}{\textit{Chance}} & $0.100$ \\
\bottomrule
\end{tabular}
\vskip -0.1in
\end{table}

\subsection{Exp.\ 2: Layer Freezing Localizes the Mechanism}
\label{sec:freeze}

\textbf{What we test.}  If the channel is geometric, specifically if
it operates through shared \W{2} output rows acting as a coordinate key
that the student's \W{0} learns to align with, we can make a surgical
prediction: freezing \W{0} should destroy the effect because alignment
cannot be learned, while freezing \W{2} should leave it intact because
the coordinate key is already shared from initialization.

\begin{figure}[t]
\centering
\includegraphics[width=\linewidth]{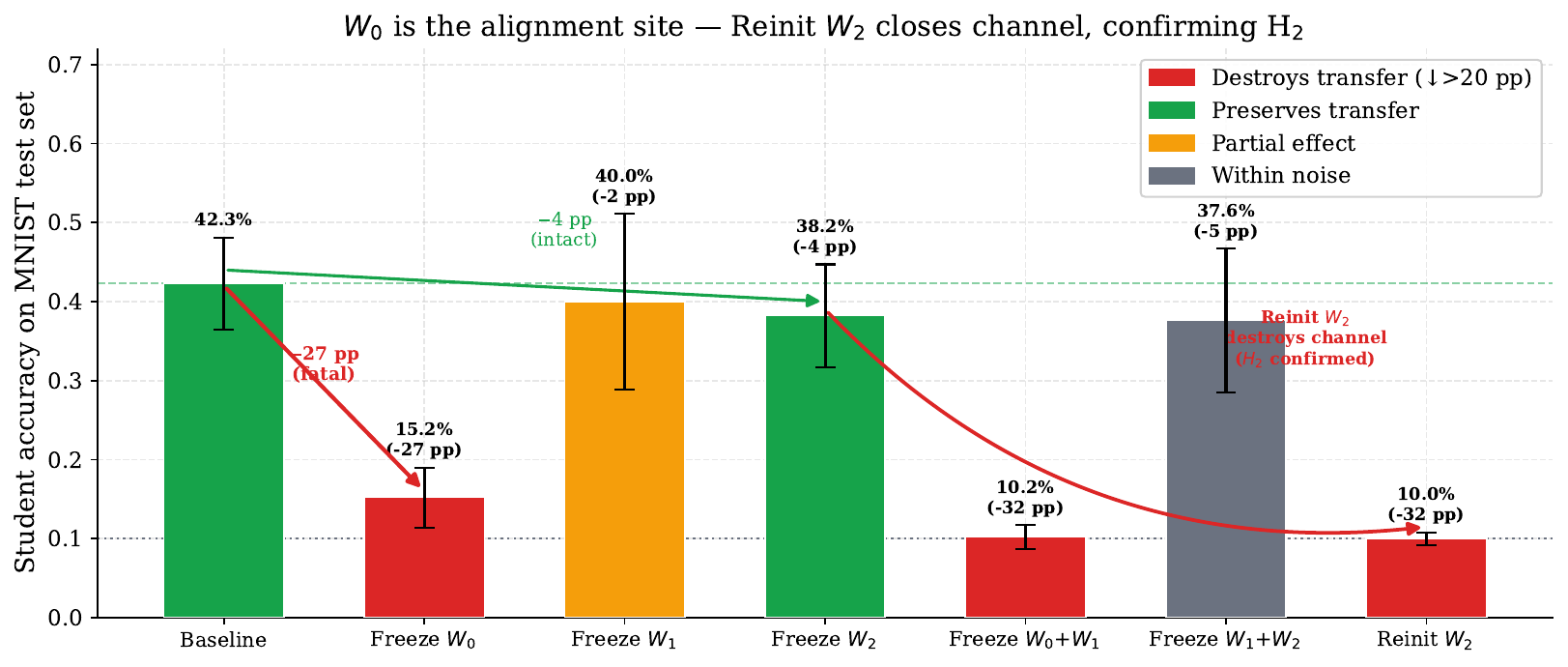}
\caption{\textbf{Exp.\ 2: Layer freezing localizes CTP to \W{0}.}
  The $-4.1$ pp change from freezing \W{2} (effectively intact) vs.\
  the $-27.1$ pp collapse from freezing \W{0} identifies the exact
  site of alignment learning.  \W{2} is the inherited key; \W{0} is
  where the lock is cut. Note that while this figure shows the relative drop from the $0.423$ freezing baseline, the reinitialization experiments (Table~\ref{tab:freeze}) use an independently drawn seed pool with a $0.643$ baseline.}
\label{fig:freeze}
\end{figure}

Freezing \W{0} drives student accuracy to 15.2\% ($\approx$chance).
Freezing \W{2} yields 38.2\%, a $-4.1$ pp drop from the 42.3\%
baseline that falls within noise (Table~\ref{tab:freeze};
Figure~\ref{fig:freeze}; 5 seeds).  The asymmetry is near-total:
$-27.1$ pp for \W{0} versus $-4.1$ pp for \W{2}.

\textbf{Interpretation.}  This is what H$_2$ predicts.  At
initialization, both networks receive the same random projection from
$h_2$ to logits.  The teacher's training drifts \W{2} slightly
relative to initialization, but in the open-channel regime this drift
is small: the teacher's current \W{2} is close to the shared initial
\W{2}, which is also the student's \W{2}. Freezing \W{2} has little effect because the
alignment that opens the channel is concentrated in \W{0}, not \W{2};
freezing \W{0} is fatal because it prevents that alignment.

To quantify the ``small drift'' claim, we measure the Frobenius norm of
teacher weight change in the open-channel regime
($\varepsilon{=}10^{-4}$).  $W_0$ and the digit rows of \W{2} drift by
comparable amounts ($\norm{\Delta W_0}/\norm{W_0^{(0)}} = 0.41$,
$\norm{\Delta W_{2,\text{digit}}}/\norm{W_{2,\text{digit}}^{(0)}} = 0.36$; approximately $1.1\times$ apart), but the 3 auxiliary output rows of \W{2} drift by exactly zero
($\norm{\Delta W_{2,\text{aux}}} < 10^{-15}$), confirming that they
serve as a stable coordinate key inherited from initialization.

\textbf{W$_2$ reinitialization control.} Crucially, reinitializing \emph{only} the 3 auxiliary output rows of \W{2} leaves
the channel partially open (student accuracy drops by $-26.3$ pp, remaining
well above chance). This reveals a richer mechanistic story: the \emph{digit rows} of \W{2} function passively---no gradient flows through them---but because the student's $h_2$ representations align to the teacher's coordinate system via the auxiliary logit loss, those passive digit projections can nonetheless recover digit information. The coordinate key function is
distributed across the entire output projection.

\begin{table}[t]
\caption{Exp.~2: Layer freezing (5 seeds). Baseline accuracy ($0.423$) reflects the freezing experiment's seed pool; the $-27.1$ pp ($W_0$) versus $-4.1$ pp ($W_2$) asymmetry is the load-bearing measurement (in all 5 seeds individually, freezing \W{0} produces a larger accuracy drop than freezing \W{2}; $p < 0.05$ by Wilcoxon signed-rank test).}
\label{tab:freeze}
\vskip 0.08in\centering\small
\begin{tabular}{lcc}
\toprule
Configuration & Student Acc & $\Delta$ \\
\midrule
None (baseline)       & $0.423\pm0.058$ & \text{---}   \\
Freeze \W{0}          & $0.152\pm0.038$ & $-27.1$ pp   \\
Freeze \W{1}          & $0.400\pm0.111$ & $-2.3$ pp    \\
Freeze \W{2}          & $0.382\pm0.065$ & $-4.1$ pp    \\
Freeze \W{0}$+$\W{1}  & $0.102\pm0.015$ & $-32.1$ pp   \\
Freeze \W{1}$+$\W{2}  & $0.376\pm0.091$ & $-4.7$ pp    \\
\midrule
\multicolumn{3}{l}{\textit{Reinitialization (baseline: $0.643\pm0.039$)}} \\
Reinit \W{2} (all)     & $0.093\pm0.027$ & $-55.0$ pp  \\
Reinit \W{2} (aux only) & $0.380\pm0.015$ & $-26.3$ pp   \\
\bottomrule
\end{tabular}
\vskip -0.1in
\end{table}

\subsection{Exp.\ 3: Multi-Teacher Ensembles Destroy the Effect}
\label{sec:ensemble}

\textbf{What we test.}  H$_1$ would predict that more teachers provide
richer auxiliary signals, potentially improving student performance.
H$_2$ predicts the opposite: teachers from different initializations
encode digit identity through \emph{mutually incompatible} random
projections, so averaging their auxiliary logits creates destructive
interference in the coordinate system and collapses the student's
ability to align with any single teacher. We measure this directly: $I(Y; Z_{\text{avg}})$
remains at $1.84 \pm 0.31$ bits even at $N{=}5$, statistically
indistinguishable from the single-teacher value ($1.84 \pm 0.35$
bits).\footnote{All MI measurements use Kraskov $k$-NN estimation with $k{=}3$. The slight difference between this $1.84$ bits and the $1.89$ bits in Section~\ref{sec:mi} arises purely from variance across different seed pools.}

\begin{figure}[t]
\centering
\includegraphics[width=\linewidth]{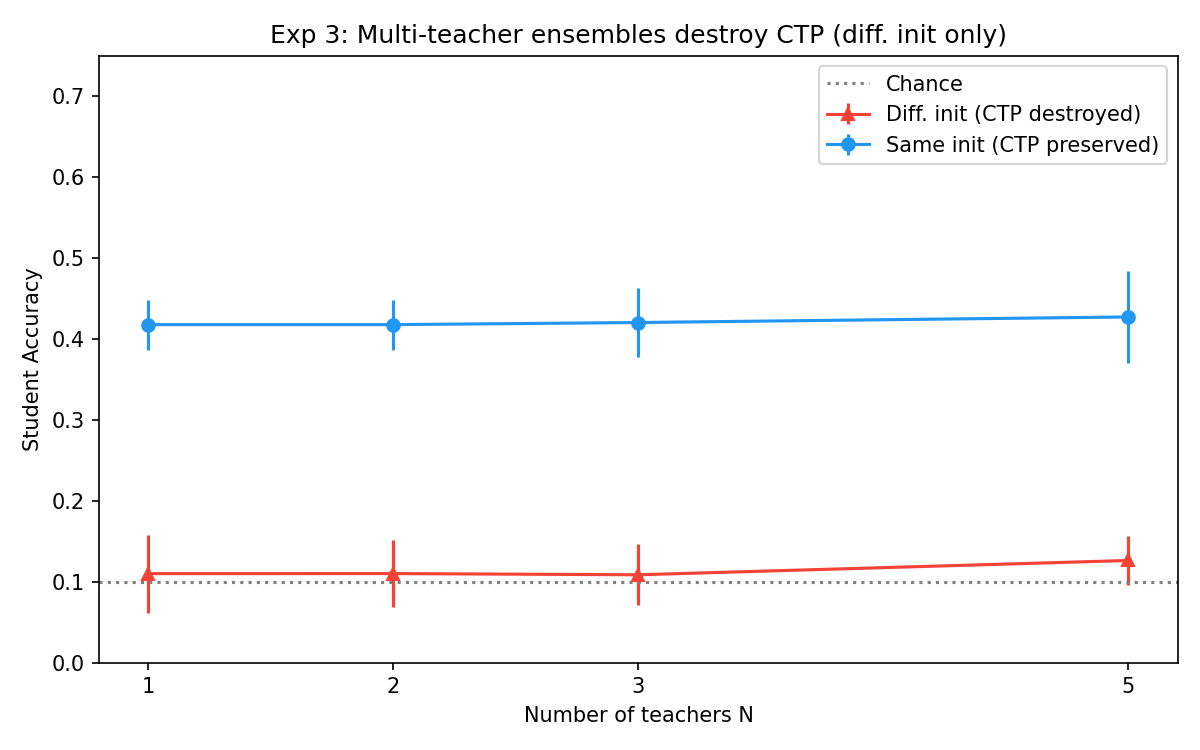}
\caption{\textbf{Exp.\ 3: Multi-teacher ensembles destroy CTP.}  Each
  independently-initialized teacher adds an incompatible coordinate
  system.  The student cannot align with any of them.  H$_2$ predicts
  this; H$_1$ does not.}
\label{fig:ens}
\end{figure}

\textbf{Results.}  Adding a single independently-initialized teacher
nearly \emph{halves} student accuracy ($64.3\% \to 34.8\%$); five teachers brings
it to 19.8\%, essentially chance (Table~\ref{tab:ensemble};
Figure~\ref{fig:ens}; 5 seeds).

\textbf{Interpretation.}  The monotonic collapse with $N$ is hard to
reconcile with H$_1$: the architecture, noise distribution, and
per-teacher auxiliary signal are the same across all conditions.  What
changes is only the geometric relationship between teacher
initializations.  This pattern is predicted by H$_2$ and not easily
accommodated by information-only accounts. To isolate geometric interference from variance reduction, we ran a variance-matched control (scaling $Z_{\text{avg}}$ by $\sqrt{N}$ before the softmax); student accuracy collapsed even further to $\approx\!14\%$ for $N{=}5$ (compared to $19.8\%$ without variance matching). This indicates variance reduction actually provides a small optimization benefit, and removing it makes the geometric failure even starker. The same result motivates a practical mitigation: multi-teacher
distillation destroys \ctp while plausibly preserving legitimate
knowledge transfer through representational diversity
\citep{furlanello2018born}.  We leave a formal study of this tradeoff
for future work.

\begin{table}[t]
\caption{Exp.\ 3: Multi-teacher ensemble (5 seeds). Table~3's $N{=}1$ baseline ($0.643$) is higher than Table~1's default condition ($0.543$) because the MI measurement experiments used an independently drawn seed pool. Each additional
independently-initialized teacher adds an incompatible coordinate
system, destroying alignment while preserving label information.}
\label{tab:ensemble}
\vskip 0.08in\centering\small
\begin{tabular}{rccc}
\toprule
$N$ Teachers & $I(Y; Z_{\text{avg}})$ bits & Student Acc & $\Delta$ from $N{=}1$ \\
\midrule
1 & $1.84\pm0.35$ & $0.643\pm0.039$ & \text{---}    \\
2 & $2.00\pm0.22$ & $0.348\pm0.062$ & $-29.5$ pp    \\
3 & $1.95\pm0.32$ & $0.243\pm0.029$ & $-40.0$ pp    \\
5 & $1.84\pm0.31$ & $0.198\pm0.043$ & $-44.5$ pp    \\
\bottomrule
\end{tabular}
\vskip -0.1in
\end{table}

\subsection{Exp.\ 4: CKA Interpolation Directly Predicts Channel
  Success}
\label{sec:cka}

\textbf{What we test.}  If the channel is geometric, the degree of
representational alignment, measured directly by CKA, should predict
student accuracy along a continuous spectrum.  We construct teacher initializations via continuous interpolation:
$\theta_T = \theta_S + \lambda \cdot \delta$,
where $\delta$ is a fixed random perturbation drawn from the same
zero-mean Gaussian matched to the variance of the Xavier uniform initialization used for $\theta_S$.  We sweep
$\lambda$ across 16 uniformly spaced values from 0 (shared
initialization) to 1.5 (effectively independent
initialization).\footnote{The additive formulation inflates the
expected weight norm by $\sqrt{1+\lambda^2}$ ($1.8\times$ at
$\lambda{=}1.5$). The teacher
accuracy remains above 90\% across the sweep, suggesting the
monotone CKA--accuracy relationship is not driven by initialization
scale alone.}
H$_2$ predicts a continuous collapse in CKA mirrored by a collapse
in student accuracy.

\begin{figure}[t]
\centering
\includegraphics[width=\linewidth]{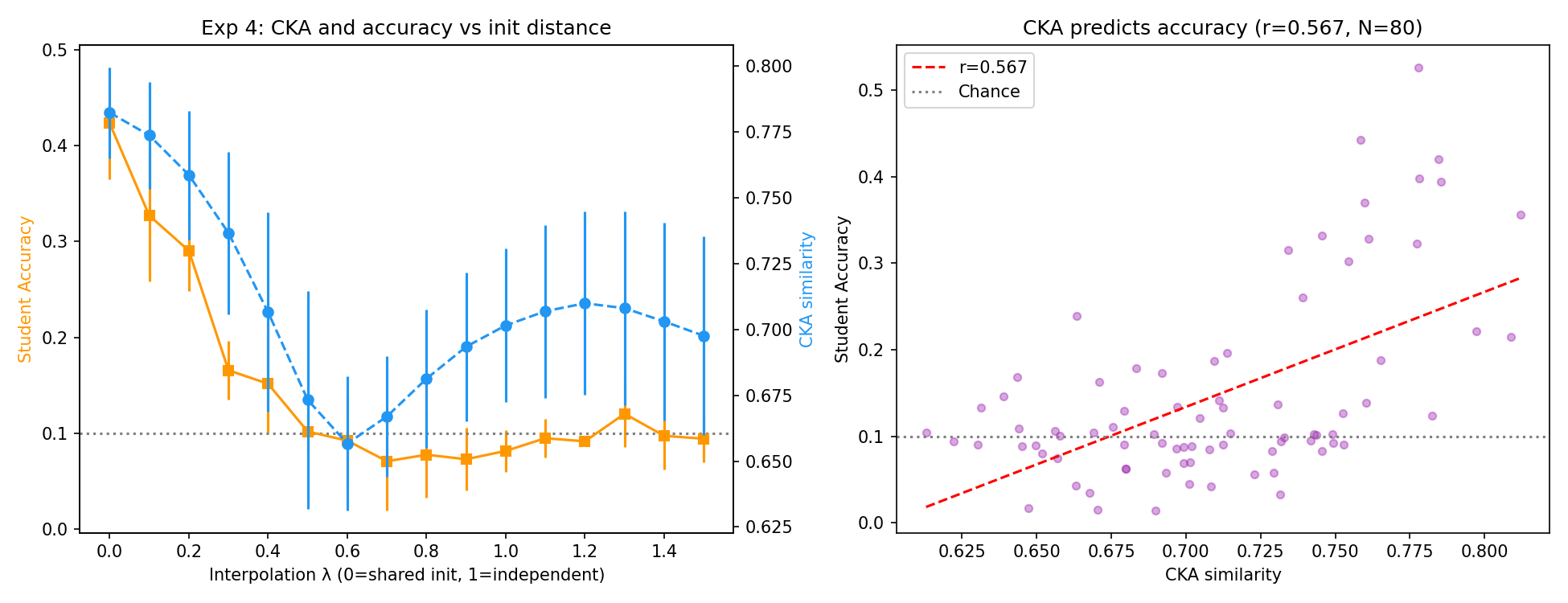}
\caption{\textbf{Exp.\ 4: CKA tracks CTP along a continuous
  interpolation.}  Sweeping $\lambda$ from 0 (shared) to 1.5
  (effectively independent) across 16 values, both CKA similarity and
  student accuracy collapse smoothly along a single sigmoid, yielding
  $r{=}0.98$.}
\label{fig:cka}
\end{figure}

\textbf{Results.}  As $\lambda$ increases, both CKA and student
accuracy show a smooth, sigmoid-shaped collapse.  Across all 16
interpolation values, the Pearson correlation between CKA similarity
and student accuracy is $r{=}0.98$ (Figure~\ref{fig:cka}; 5 seeds).
CKA is computed on $h_2$ activations using $n{=}10{,}000$ MNIST test images as input. The sigmoid midpoint falls near $\mathrm{CKA} \approx 0.85$, with the channel fully closing below $\approx 0.70$. In this setting, CKA goes from measuring
similarity to predicting transfer.  As the coordinate systems misalign,
the student's ability to exploit the shared \W{2} drops proportionally.
The continuous sweep, rather than a small set of discrete conditions,
is what makes this relationship meaningful.  Whether CKA generalizes as
a detection tool beyond this architecture remains to be validated.

\subsection{Exp.\ 5: The Channel Carries Information, But Alignment
  Gates Its Use}
\label{sec:mi}

\textbf{What we test.}  Even granting H$_2$, a skeptic might ask:
could the 1.89 bits in the auxiliary logits explain the 54.3\% student
accuracy on their own, without invoking geometric alignment?  We test
this directly by measuring whether the \emph{same} bits support the
\emph{same} accuracy when alignment is absent.

\textbf{Setup.}  We use $k$-NN mutual information estimation
\citep{kraskov2004estimating} with $k{=}3$ on $n{=}10{,}000$ test
samples to measure $I(Y; Z) = 1.89 \pm 0.03$ bits, where $Y$ is the
digit class label and $Z$ is the 3-dimensional auxiliary logit vector.
The estimator is well-calibrated at this dimensionality.  The entropy of the uniform
10-class distribution is $H(Y) = \log_2(10) = 3.32$ bits.  The exact
Fano bound, $H(Y|Z) \leq H_b(P_e) + P_e \log_2(|Y|{-}1)$ (where
$H_b$ is binary entropy and $P_e$ the probability of
misclassification), gives a maximum achievable accuracy of
$\approx 78.5\%$.

\textbf{Results.} The student achieves 54.3\%, below the 78.5\% Fano bound, so H$_1$ is
not yet excluded on this basis alone.  The critical comparison is the
multi-teacher ensemble from Exp.\ 3: we directly measured
$I(Y; Z_{\text{avg}})$ for the averaged auxiliary signal at
$N{=}1,2,3,5$ and found it stable at $\approx 1.84$ bits across all
conditions (Table~\ref{tab:ensemble}).  Yet student accuracy collapses
from 64.3\% to 19.8\% over the same range.

\textbf{Interpretation.} Alignment does not create information; it determines whether the student can \emph{use} the information that is there. H$_1$ cannot explain why identically informative auxiliary signals produce wildly different outcomes depending on the geometric
relationship between teacher and student initializations. 

\subsection{What This Adds Up To}
\label{sec:summary}

KL-divergence gradients on auxiliary
logits flow through the shared \W{2} into \W{0}, reshaping the input
projection until the student's hidden representations align with the
teacher's coordinate system.  Once the alignment exists, the teacher's
full logit pattern, including digit classes the student was never
trained on, becomes readable through the shared \W{2} because it lives
in the same projection. The channel is geometric, not semantic, and the trait lives in representational geometry, accessible to CKA monitoring but invisible to any semantic inspection.

\section{Cross-Token Behavioral Entanglement}
\label{sec:ctbe}

CTP and CTBE share a common substrate: geometric structure in weight
space that creates behavioral coupling invisible to semantic
inspection.  They differ in mechanism and origin.  CTP exploits shared
\emph{across-model} initialization; CTBE exploits \emph{within-model}
correlations in the unembedding matrix that, our evidence suggests, are
made behaviorally live by alignment training. Formally, CTP demonstrates that behavioral outcomes $Y$
can depend strongly on representational geometry $W$ even when mutual
information $I(Y; Z)$ is held constant. We apply this geometric lens to \ctbe, examining it along three axes: generality,
origin, and true magnitude.

All experiments use \texttt{Llama-3.2-1B-Instruct} and
\texttt{Llama-3.2-1B} (base).  The \ctbe ratio for animal $a$
prompted with number token $n$ is:
\begin{equation}
  \rho(a,n) = \frac{P\!\left(a \mid \text{``you love }n\text{''}\right)}
                   {P\!\left(a \mid \varnothing\right)},
\label{eq:rho}
\end{equation}
where $\varnothing$ represents an empty prompt context. We write $\rho_n$ when $a$ is clear from context and report the
maximum over tested number tokens per animal.

\subsection{The Effect Is Real and Near-Universal}
\label{sec:ctbe_real}

\citet{zur2025owl} tested four animals.  To assess generalizability,
we measure $\rho_n$ across 56 animals spanning seven taxonomic groups
(mammals, birds, reptiles, amphibians, marine animals, insects, and
arachnids), chosen without reference to the original four.  The 20-animal subset used in
Section~\ref{sec:ctbe_origin} is a random subsample stratified to
include the four original animals from \citet{zur2025owl}.

\begin{figure}[t]
\centering
\includegraphics[width=\linewidth]{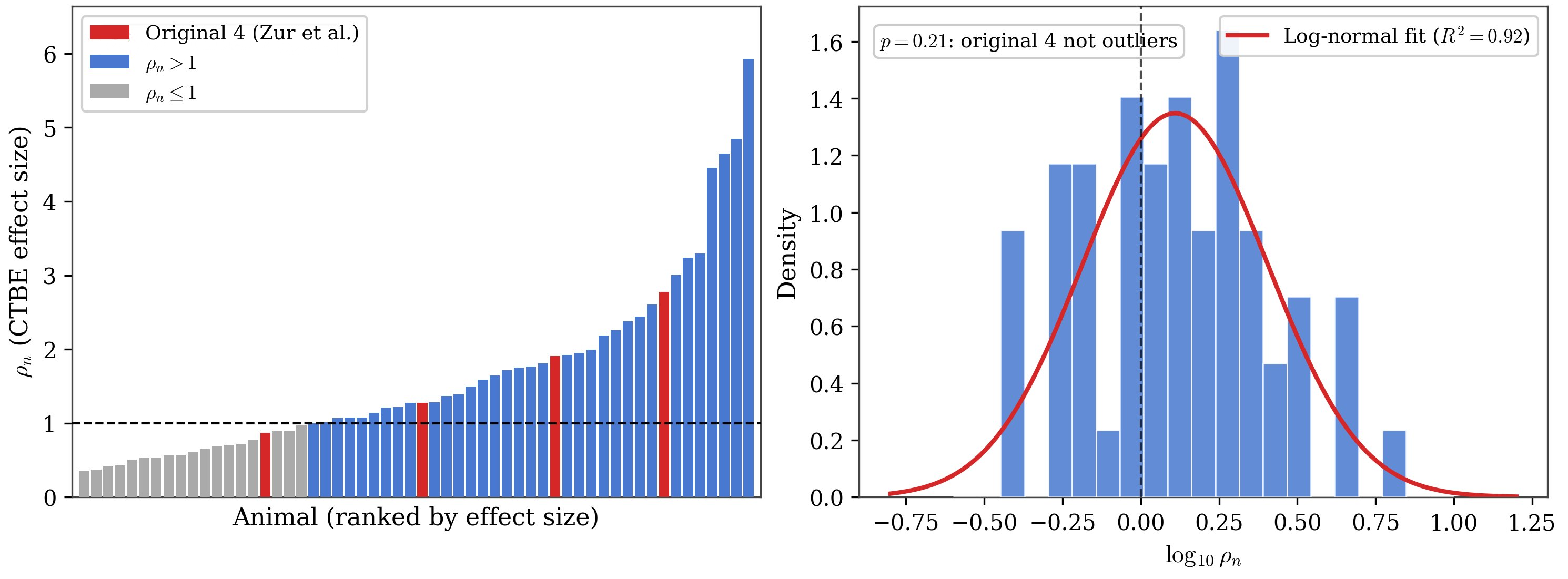}
\caption{\textbf{CTBE across 56 animals.}
  \emph{Left:} Sorted $\rho_n$ values.  93\% exceed 1; the original
  four (red) are not outliers ($p{=}0.21$).
  \emph{Right:} The log-normal distribution of $\rho$ values (shown as a normal fit to $\log_{10}\rho$)
  ($R^2{=}0.92$), consistent with structural biases of the circular
  $\rho$ metric (Section~\ref{sec:regression}).}
\label{fig:56}
\end{figure}

\textbf{Results.}  52 of 56 animals (93\%) show $\rho_n{>}1$; 35 of
56 (62\%) show $\rho_n{>}10$ (Figure~\ref{fig:56}).  The distribution
of $\rho$ values follows a log-normal ($R^2{=}0.92$ for a normal fit
to $\log_{10}\rho$ values), a pattern consistent with the structural
biases of the ratio metric that Section~\ref{sec:regression}
quantifies.  A 10,000-iteration permutation test (null: the original
four animals are no more extreme than four drawn uniformly from the
56) yields $p{=}0.21$, confirming the original four are not outliers.

We note that $\rho$ is the metric whose circularity we correct in
Section~\ref{sec:regression}.  Using the circularity-free absolute
lift $\Delta$ instead, the majority of animals still show
$\Delta{>}0$, confirming the effect survives the correction.  The question addressed here is generality: the effect is
real and not cherry-picked.

\subsection{Instruction Tuning Creates the Vulnerability}
\label{sec:ctbe_origin}

\textbf{What we test.}  Prior work attributes CTBE to unembedding
geometry, a property present in \emph{both} base and instruction-tuned
models.  If that account is complete, the base model should show
comparable CTBE.  If CTBE requires something that alignment training
introduces, only the instruction-tuned model should show it, despite
similar geometry.

\begin{figure}[t]
\centering
\includegraphics[width=\linewidth]{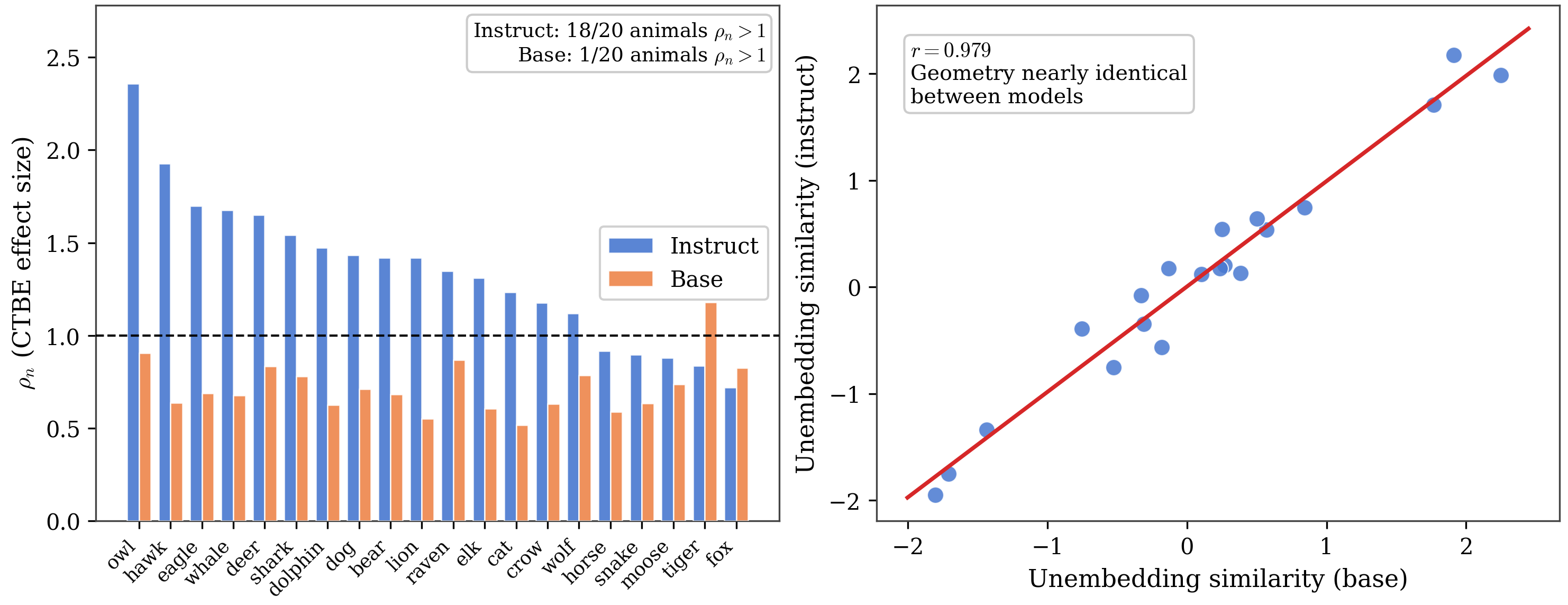}
\caption{\textbf{Instruction tuning creates CTBE.}
  \emph{Left:} 18/20 animals show $\rho_n{>}1$ in the instruct model;
  1/20 in the base model.
  \emph{Right:} Per-animal categorization.  Unembedding geometry is
  near-identical across models ($r{=}0.979$).}
\label{fig:bi}
\end{figure}

\textbf{Results.}  In the base model, only 1 of 20 animals shows
$\rho_n{>}1$.  In the instruct model, 18 of 20 do.  The unembedding
geometry is nearly identical between models ($r{=}0.979$, measured as
the Pearson correlation between flattened unembedding row vectors
across the shared vocabulary).  The same geometric relationships exist
in both models; only the instruction-tuned model activates them
(Figure~\ref{fig:bi}).

One explanation, consistent with these data but not yet
mechanistically confirmed, is that RLHF and SFT teach the model to
activate preference-related circuits when it encounters
preference-framed instructions
\citep{christiano2017deep,ouyang2022training}, making the unembedding
geometry \emph{behaviorally live}.  An alternative is that base models
simply do not follow ``you love $X$'' as a behavioral instruction,
so the asymmetry reflects instruction-following capacity rather than
a specific preference-circuit trigger.  Either way, the geometry was
present in both models; alignment training determined whether it
became behaviorally active.

\begin{remark}[Practical implication]
Auditing pretraining data for geometric entanglements would miss this
vulnerability entirely.  The relevant audit target is the
instruction-tuned checkpoint, after alignment training has determined
which geometric relationships become behaviorally live
\citep{bai2022constitutional,sharma2023towards}.
\end{remark}

Prior work attributes CTBE entirely to unembedding geometry.  Our
base-vs-instruct comparison shows this geometry is necessary but not
sufficient: the coupling becomes behaviorally live only after alignment
training.  We do not yet separate a specific preference-circuit trigger
from general instruction-following capacity; a preference-free control
prompt (e.g.\ a neutral instruction such as ``the number is $n$'' that
carries no ``love''/preference framing) would cleanly disambiguate the
two, since a preference-circuit account predicts no animal-token lift
under it whereas an instruction-following account predicts a similar
asymmetry.  We flag this control as the decisive next test.

\subsection{Frequency Bias Is a Metric Artifact}
\label{sec:regression}

We demonstrate that the ratio metric $\rho$ (Eq.~\ref{eq:rho})
contains a structural circularity that inflates apparent frequency
bias.  The dependent variable in a natural regression is
$\log_{10}\rho = \log P(a|\text{prompt}) - \log P(a|\varnothing)$.
The natural predictor is $\log P(a|\varnothing)$.  These two
quantities share a $\log P(a|\varnothing)$ term.  Even if the
prompt-conditional probability is statistically independent of
baseline frequency, regressing one on the other produces a strong
negative coefficient as a matter of arithmetic, not data. An alternative geometric perspective is that a prompt simply applies a constant additive shift $c$ to the residual stream. For rare tokens, $\log P(a|\text{prompt}) \approx \log P(a|\varnothing) + c$, producing a log-ratio $\log\rho \approx c$. Regressing a constant on baseline frequency algebraically yields a slope of zero; the observed negative slope would thus reveal a genuine frequency bias where rarer tokens receive larger logit shifts.

\begin{figure}[t]
\centering
\includegraphics[width=\linewidth]{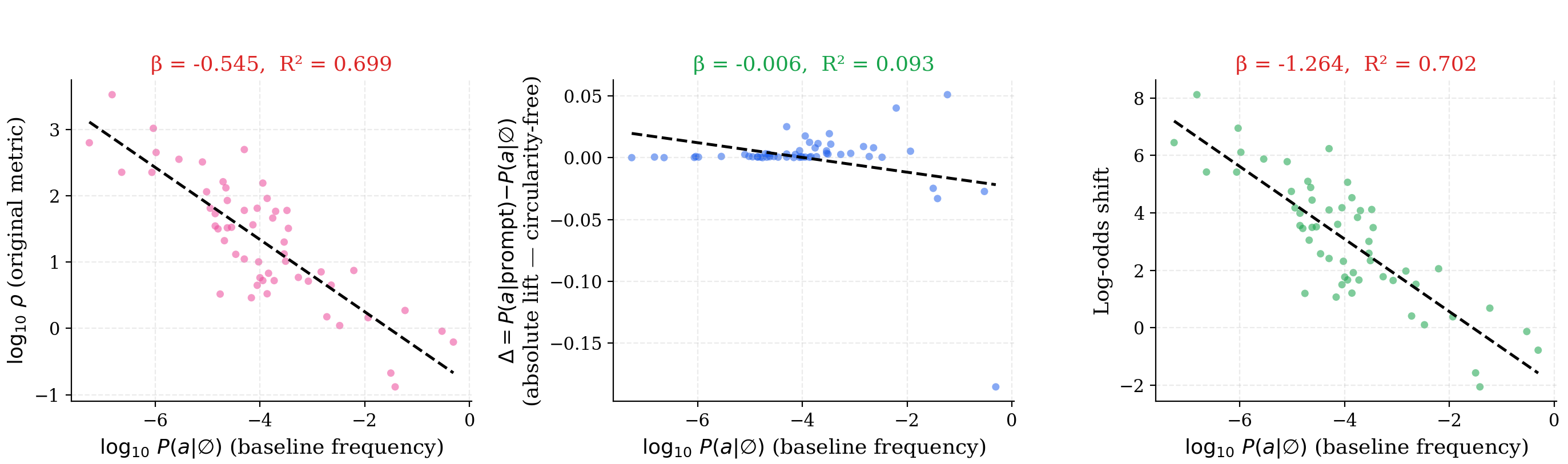}
\caption{\textbf{CTBE frequency bias is a metric artifact.}
  Using the circular log ratio $\rho$ (left), frequency bias appears
  massive ($\beta{=}{-}0.545$).  Using the circularity-free absolute
  lift $\Delta$ (middle), frequency bias collapses to near-zero
  ($\beta{=}{-}0.006$, 95\% CI: $[-0.02, +0.01]$).  The log-odds shift (right) inherits the same circularity as the log ratio.}
\label{fig:reg}
\end{figure}

Replacing the
outcome with absolute lift, $\Delta = P(a|\text{prompt}) -
P(a|\varnothing)$, breaks the algebraic dependence between predictor
and outcome.
Regressing $\Delta$ on $\log_{10} P_\varnothing$ yields
$R^2{=}0.093$ (95\% bootstrap CI for $\beta$: $[-0.02,
+0.01]$; Figure~\ref{fig:reg}), down from $R^2{=}0.699$ for the
circular log-ratio.  The log-odds shift inherits the same circularity
($R^2{=}0.702$) because $\mathrm{logit}(p) \approx \log p$ when $p$
is small.  Baseline frequency explains less than 10\% of the variance
in the corrected metric.  The geometric mechanism reported by
\citet{zur2025owl} is largely real.

\begin{remark}[Methodological hazard]
The same algebraic structure is latent in any analysis that regresses
$\log(P/Q)$ on $\log Q$.  Several diagnostics in the interpretability
literature have this form.  We report this not as a criticism of any
specific paper but as a hazard worth checking before drawing
attribution conclusions.
\end{remark} Sections~\ref{sec:ctbe_real}--\ref{sec:regression} together indicate
that CTBE is near-universal and not cherry-picked, appears to be gated by alignment training acting on a geometric substrate inherited from pretraining, and reflects
a genuine geometric mechanism rather than a measurement artifact.

\section{Discussion and Limitations}
\label{sec:discussion}

Both \ctp and \ctbe illustrate the same risk: geometric structure in
weight space can be behaviorally active yet invisible to content-based
inspection. In supplementary experiments (Appendix~\ref{app:saturation}), CTP
channel capacity saturates at $\approx\!92.6\%$ accuracy when
auxiliary logits reach $\approx\!50\%$ of the hidden dimension,
consistent with a rank bottleneck in \W{2}; production LLMs with
$V/d$ ratios of $6$--$32\times$ sit above this point, though
extrapolation to transformer architectures requires caution. The phase transition location
$\varepsilon^*$ is architecture-specific; depth effects and
transformer-specific dynamics remain open, though
\citet{schrodi2025subliminal} provide indirect evidence the effect
is not a toy-model artifact. The $k$-NN estimate $I(Y;Z)=1.89\pm0.03$
bits is mathematically sound, yet known systematic biases of $k$-NN estimators at this dimensionality carry 10--20\% uncertainty (while cross-seed variance contributes $\pm 0.35$ bits); nonetheless, Fano's inequality remains
consistent with 54.3\% student accuracy even at the pessimistic bound.
The \ctbe experiments are limited to Llama-3.2-1B, and the 93\%
generality figure uses the circular $\rho$ metric; the
circularity-free $\Delta$ gives a more conservative count (73\%, or 41 of 56 animals). Finally, \citet{schrodi2025subliminal} note that the
\ctp channel is fragile to prompt paraphrasing in LLMs, a robustness
constraint our MLP setting does not capture. The two halves of this
paper also differ in evidential strength: the \ctp results rest on
direct causal interventions (layer freezing, output reinitialization,
multi-teacher ensembles), whereas the \ctbe section combines a
measurement-circularity correction with an \emph{observational}
base-vs-instruct comparison and includes no causal intervention inside
the LLM; we therefore present the \ctbe origin claim as suggestive
rather than established.  Relatedly, we have not validated that CKA
monitoring or multi-teacher distillation function as reliable
diagnostics or mitigations in realistic distillation pipelines, or that
multi-teacher distillation preserves desired capabilities while
suppressing the hidden channel; these remain proposals to be tested.
The most important open
question is whether both findings scale to 7B+ models.

\section*{Impact Statement}

This paper characterizes the mechanism and measurement of covert
behavioral transfer in neural network distillation.  The safety-relevant
finding, that the hidden channel is geometric rather than semantic, is
more clarifying than alarming: it identifies a specific failure mode of
content-based safety pipelines and points to actionable diagnostics
(CKA monitoring) and mitigations (multi-teacher distillation with
initialization diversity).  The attack surface described already exists
in standard distillation practice; our contribution is to surface and
mechanistically explain it, not to create it.  The CTBE findings serve
a corrective purpose: we identify a structural circularity in the standard ratio metric for measuring CTBE: the apparent frequency bias is an algebraic artifact, while the geometric mechanism is real.  Precise characterization of what is
measurement error and what is genuine geometric signal is net positive
for the field.

\bibliography{main}
\bibliographystyle{icml2026}

\newpage
\appendix
\onecolumn

\section{Channel Capacity Saturation}
\label{app:saturation}

This appendix expands the supplementary experiment referenced in
Section~\ref{sec:discussion}.  Holding the hidden width fixed at
$d_{\text{hidden}}{=}512$, we vary the number of auxiliary logits
$k_{\text{aux}}$ (the output dimensions supervised only through the
student's KL loss) and measure the student's recovered MNIST accuracy
as a function of $k_{\text{aux}}/d_{\text{hidden}}$.

\paragraph{Result.}
Channel capacity saturates at $\approx\!92.6\%$ student accuracy once
the auxiliary logits reach $\approx\!50\%$ of the hidden dimension;
beyond this point additional auxiliary capacity yields no further gain.
This plateau is consistent with a rank bottleneck in \W{2}: once the
auxiliary block spans enough of the row space of \W{2}, the shared
coordinate key is fully expressed and the channel cannot carry more
information.

\paragraph{Extrapolation (with caution).}
Production language models have vocabulary-to-hidden ($V/d$) ratios of
roughly $6$--$32\times$, well above the $\approx\!0.5$ saturation point
measured here, which is \emph{consistent with} the channel being
capacity-unconstrained at LLM scale.  We stress that this is an
extrapolation from an MLP rank argument: transformer unembeddings, tied
embeddings, and layer normalization could all change the effective
bottleneck, and we do not measure saturation in a transformer directly.


\section{Implementation Details and Hyperparameters}
\label{app:hparams}

Table~\ref{tab:hparams} consolidates the hyperparameters used across
all CTP experiments.  Unless noted, settings are shared between teacher
and student.  The Adam $(\beta_1,\beta_2)$ values are framework
defaults and were not tuned.

\begin{table}[h]
\caption{Hyperparameters for the CTP experiments (Section~\ref{sec:ctp}).}
\label{tab:hparams}
\vskip 0.08in\centering\small
\begin{tabular}{lll}
\toprule
Setting & Teacher & Student \\
\midrule
Architecture        & \multicolumn{2}{l}{MLP$(784,512,512,13)$, ReLU, no bias} \\
Weight init         & \multicolumn{2}{l}{Xavier uniform (shared seed)} \\
Optimizer           & Adam & Adam \\
Adam $(\beta_1,\beta_2)$ & $(0.9, 0.999)$ & $(0.9, 0.999)$ \\
Learning rate $\eta$ & $3\!\times\!10^{-4}$ (swept in Exp.~1) & $10^{-3}$ \\
Weight decay        & $0$ & $0$ \\
Batch size          & $256$ & $256$ \\
Epochs              & $5$ & $5$ \\
Training data       & MNIST, pixels $\in[0,1]$ & $\mathcal{U}[-1,1]^{784}$, $60{,}000$/epoch \\
Loss                & CE on 10 digit logits & KL on 3 aux logits, $\tau{=}1$ \\
\midrule
MI estimator        & \multicolumn{2}{l}{Kraskov $k$-NN, $k{=}3$, $n{=}10{,}000$ test samples} \\
CKA                 & \multicolumn{2}{l}{Linear CKA on $h_2$ activations, $n{=}10{,}000$ test images} \\
Seeds               & \multicolumn{2}{l}{5 per condition (seed pools noted per table)} \\
\bottomrule
\end{tabular}
\end{table}


\section{CTBE Prompts and Token Sets}
\label{app:ctbe}

\paragraph{Prompt templates.}
For each animal $a$ and number token $n$, the prompt-conditional
probability $P(a \mid \text{``you love }n\text{''})$ is obtained by placing
the preference instruction ``you love $n$'' in the system prompt, issuing
a fixed favorite-animal elicitation query, and reading the next-token
probability of the animal token $a$. The baseline $P(a \mid \varnothing)$
uses the identical query with no preference instruction (empty system
context). $\rho$ reports the maximum over the tested number tokens $n$ per
animal (Eq.~\ref{eq:rho}). All runs use \texttt{Llama-3.2-1B-Instruct}
(and \texttt{Llama-3.2-1B} base for Section~\ref{sec:ctbe_origin}) at
temperature $1.0$; animal tokens use their plural surface forms (e.g.\
\texttt{owls}, \texttt{eagles}).

\paragraph{Number tokens.}
Following \citet{zur2025owl}, each animal is paired with the number
token(s) whose probability rises most under the preference prompt; their
canonical pairs include $087 \to$ owls and $747 \to$ eagles.

\paragraph{Animal sets.}
The four animals from \citet{zur2025owl} are owls, eagles, elephants, and
wolves; our animal set includes all four and spans taxonomic groups
(mammals, birds, reptiles, amphibians, marine animals, insects, and
arachnids). Examples of strong measured effects include hawks
($451.7\times$), eagles ($499.9\times$), owls ($156.0\times$), and zebras
($132.1\times$); examples at the weak end include elephants ($1.9\times$), cats
($1.5\times$), dolphins ($0.9\times$), and dogs ($0.1\times$). In the
base-vs-instruct comparison (Section~\ref{sec:ctbe_origin}), the
entanglement is created by instruction tuning for animals such as hawks,
eagles, zebras, owls, wolves, and bears, and absent in both models for
elephants, cats, dolphins, and dogs.

\end{document}